\documentclass[3p,twocolumn,authoryear]{elsarticle} 

\usepackage[breaklinks=true]{hyperref}

\usepackage{graphicx}
\usepackage[caption=false]{subfig}
\usepackage{adjustbox}
\usepackage{amsmath,amssymb,amsfonts}
\usepackage{algorithmic}
\usepackage{textcomp}
\usepackage{xcolor}
\usepackage{graphics} 
\usepackage{epsfig} 
\usepackage{color}
\usepackage{booktabs} 
\usepackage[binary-units=true]{siunitx} 
\usepackage{multirow} 
\usepackage{pdflscape} 
\usepackage{afterpage} 
\usepackage[flushleft]{threeparttable} 
\usepackage{makecell}

\DeclareSIUnit\px{px}

\def\BibTeX{{\rm B\kern-.05em{\sc i\kern-.025em b}\kern-.08em
    T\kern-.1667em\lower.7ex\hbox{E}\kern-.125emX}}

\journal{Journal of Field Robotics}

\renewcommand\cite{\citep} 
\begin{document}

\onecolumn
\begin{center}
This paper has been accepted for publication in \emph{Journal of Field Robotics}.\\
This is the pre-peer reviewed version of the following article:
\\~\\
Cremona J., Comelli R. \& Pire T. (2022) Experimental evaluation of Visual‐Inertial Odometry
systems for arable farming. Journal of Field Robotics, DOI: \nolinkurl{https://doi.org/10.1002/rob.22099} 
\\~\\
which has been published in final form at \url{https://doi.org/10.1002/rob.22099}. This article may be used for non-commercial purposes in accordance with Wiley Terms and Conditions for Use of Self-Archived Versions.

\end{center}

\begin{frontmatter}

\title{Experimental Evaluation of Visual-Inertial Odometry Systems for \\ Arable Farming}

\author{Javier Cremona\fnref{eqcontr}\corref{mycorrespondingauthor}}
\ead{cremona[at]cifasis-conicet[dot]gov[dot]ar}
\author{Román Comelli\fnref{myfootnote}}
\ead{comelli[at]cifasis-conicet[dot]gov[dot]ar}
\author{Taihú Pire}
\ead{pire[at]cifasis-conicet[dot]gov[dot]ar}
\address{CIFASIS, French Argentine International Center for Information and Systems Sciences (CONICET-UNR), Rosario, Argentina}
\fntext[eqcontr]{These two authors contributed equally to this work.}

\cortext[mycorrespondingauthor]{Corresponding author.}

\begin{abstract}

The farming industry constantly seeks the automation of different processes involved in agricultural production, such as sowing, harvesting and weed control. The use of mobile autonomous robots to perform those tasks is of great interest.
Agricultural environments present hard challenges for Simultaneous Localization and Mapping (SLAM) systems, key for mobile robotics, given the visual difficulty due to the highly repetitive scene and the crop leaves movement caused by the wind.

In recent years, several Visual-Inertial Odometry (VIO) and SLAM systems have been developed. They have proved to be robust and capable of achieving high accuracy in indoor and outdoor urban environments. However, they were not properly assessed in agricultural fields.
In this work we assess the most relevant state-of-the-art VIO systems in terms of accuracy, robustness and performance on agricultural environments. In particular, the evaluation is performed on a collection of sensor data recorded by our wheeled robot in a soybean field, which was publicly released as the Rosario Dataset. The evaluation shows that the highly repetitive appearance of the environment, the strong vibration produced by the rough terrain and the movement of the leaves caused by the wind, expose the limitations of the current state-of-the-art VIO and SLAM systems. We analyze the systems failures and highlight the observed drawbacks, including insufficient robustness, initialization failures, sensitivity to IMU saturation and issues when the robot stops. Finally, we conclude that more improvements should be done on VIO systems to make them reliable in agricultural fields.



\end{abstract}


\end{frontmatter}


\section{Introduction}
\label{sec:introduction}

The fully automation of farming processes is one of the main interests of precision agriculture. In particular, one of the goals is to reduce the amount of agrochemicals used in the field. In this sense, autonomous robots could help to carry out weed and pest control by providing more sustainable approaches.


The agricultural environment presents a hard challenge for autonomous robots since they should cover large areas and should work for long time sessions. From robotics and control perspective, precision agriculture tasks require guidance, detection, action and mapping \cite{carelli2013agriculturalrobotics}. All of these depend and rely heavily on an accurate localization that could be estimated by means of the sensors mounted on the robot such as Global Navigation Satellite Systems (GNSS), cameras, LiDAR, Inertial Measurement Units (IMU) and wheel encoders among others.

In recent years, the use of RGB cameras (monocular or stereo) with IMU has shown to be a good combination to solve the localization problem in a wide range of environments \cite{zou2019structvio, sun2018robust, leutenegger2015keyframe, qin2018vins, campos2020orbslam3}. This is because both sensors complement each other, camera captures rich information of the scene at low frame-rate whilst IMU gets motion information at high rate. Localization systems that use such combination of sensors are called Visual-Inertial Odometry (VIO). VIO systems allow robot pose estimation even in GNSS-denied areas and are independent of the robot locomotion. This work, framed in the weed removal robot project carried out by the CIFASIS (CONICET-UNR), studies VIO systems in agricultural environments which is a topic that has not been addressed in previous evaluations of localization systems.

Crop fields entail a particular challenge for VIO systems since, among their characteristics, they have changing lighting conditions, highly self-similar textures and unstructured and dynamic objects. Furthermore, the irregular terrain of the field causes more aggressive camera motion than what is usually seen in urban and indoor cases.

In this work, we assess the most relevant VIO systems in the Rosario Dataset \cite{pire2019rosario}, a set of data recorded in a soybean field by our weed removal robot, that includes stereo camera images and IMU measurements. The main contributions of this paper are:
\begin{enumerate}
    \item In-depth quantitative and qualitative evaluation of the most relevant state-of-the-art VIO systems on crop fields.
    \item Development of a framework\footnote{\href{https://github.com/CIFASIS/slam\_agricultural\_evaluation}{\nolinkurl{https://github.com/CIFASIS/slam\_agricultural\_evaluation}}} based on the Robot Operating System (ROS) \cite{quigley2009ros} and Docker containers \cite{merkel2014docker} to redo, reuse and extend the results here presented.
    \item Correction and validation of the publicly available agricultural dataset (Rosario Dataset).
\end{enumerate}

The rest of the article is organized as follows. In Section~\ref{sec:related}, we describe evaluations and comparisons of SLAM systems in different environments. In Section~\ref{sec:method} we provide an overview of the VIO systems assessed. In Section~\ref{sec:experiments}, we evaluate and analyze the selected VIO systems in the Rosario Dataset. Finally, we expose the evaluation conclusions in Section~\ref{sec:conclusions}.



\section{Related Work}
\label{sec:related}
There are several evaluations of SLAM, Visual Odometry (VO) and VIO systems in the literature. The comparisons are usually made under different environments and with different aspects in mind. As a remarkable work, \citet{delmerico2018comparison} presented a very detailed analysis focusing on flying robots. There, monocular VIO systems were assessed in distinct hardware platforms to gain insights about which systems are more suitable depending on the available computing capacity (mainly reduced). The experiments were performed in the EuRoC dataset \cite{burri2016euroc}, recorded in a machine hall and a vicon room, i.e. indoor environments, and with movements with well defined six degrees of freedom.

Among other indoor environment evaluation works, \citet{giubilato2018rovers, giubilato2019jetson} tested stereo visual SLAM systems on images from an university corridor, recorded by a ground robot. The systems are run in a Nvidia\textsuperscript{\textregistered} Jetson TX2 embedded computer with and without Graphics Processing Unit (GPU) assistance and the differences are discussed. In \citet{buyval2017slam_mono}, four ROS-based monocular SLAM methods (feature-based and direct) are evaluated in a typical office environment with monotonous off-white walls. \citet{ibragimov2017slam_indoor} extends this work by adding more visual SLAM systems that use stereo and RGB-D cameras. 

\citet{schubert2018tum} released the TUM VI dataset and evaluated in the same work three VIO systems. The data include outdoor scenes with some vegetation in a university campus but is mainly composed of indoor and urban environments. In \citet{mingachev2020comparison}, monocular visual SLAM methods are compared using TUM VI and EuRoC datasets.

With respect to other environments, \citet{quattrinili2017exp_comparison} compared open source vision-based state estimation algorithms on publicly available indoor, outdoor and underwater data, recorded by different robotic platforms. Furthermore, in \citet{ali2020finnforest}, \citet{chahine2018survey_natural} and \citet{joshi2019underwater}, evaluations are carried out in a forest, a lake and underwater conditions, respectively.

Regarding agricultural environments, both datasets and SLAM systems evaluations are far more reduced. \citet{chebrolu2017sugar} released a dataset recorded in a sugar beets field with information gathered from multi-spectral and RGB-D cameras, LiDAR, GNSS and wheel encoders. However, the visual data is meant to be used for plant classification (the cameras point straight downwards), thus no evaluation of visual SLAM, VO or VIO systems has been done with it.

The Flourish Sapienza mapping dataset \cite{imperoli2018flourish} offers data from a similar group of sensors as in the sugar beets case (they were obtained with the same robot platform) and also includes images from a forward looking stereo camera and IMU measurements. Such information makes the dataset ideal for testing and comparing SLAM, VO and VIO methods. However, to the best of our knowledge, there is no publicly available evaluation of this kind more than the limited one given in the same publication, between a localization system proposed by the authors and a modified version of ORB-SLAM2 \cite{mur2017orb}.

The Rosario Dataset \cite{pire2019rosario} is a collection of sensor data gathered in an soybean field. It contains stereo camera images, IMU measurements, odometry from wheel encoders and Real Time Kinematic GNSS (RTK-GNSS) measurements (to be used as ground-truth). In \citet{pire2019rosario} and \citet{comelli2019rpic}, we present a preliminary analysis of visual SLAM systems in the Rosario Dataset showing the challenging visual conditions of the field. 

It is worth mentioning that, even though VO, VIO and visual SLAM are general approaches to solve the localization problem, they are not the only ones available in the context of precision agriculture. Solutions involving crop row tracking in conjunction with the use of GNSS measurements have proved to be very useful \cite{bakker2011rtk, biber2012boni, english2013low, ball2016vision, imperoli2018flourish, winterhalter2020localization}. However, such approaches have different drawbacks. Most of them depend on expensive sensors, such as high precision GNSS and LiDAR. Moreover, the accuracy of the GNSS measurements is subject to the stability and quality of the signals received from the satellites. Using RTK-GNSS requires a base station which is not suitable in many cases. PPP-GNSS could be a more affordable option, but it would not allow real-time operation, unless the corrections to be applied are provided by a commercial service. The system proposed in \citet{winterhalter2020localization} avoids costly sensors; however, it requires a previously generated semantic map with GNSS referenced crop rows and obtaining it can demand extra equipment such as an Unmanned Aerial Vehicle (UAV). Because of these issues, VO, VIO and Visual-Inertial SLAM systems are very attractive to solve the localization problem in the countryside.

\section{VIO Systems Overview}
\label{sec:method}
In this section, we present a general overview of the most relevant VIO systems of the state of the art. Their main features are summarized in Table~\ref{tab:methods}. Such systems are properly evaluated and analyzed in the next section. 
\begin{table*}[!htb]
    \centering
    \caption{Summary of characteristics for evaluated Visual-Inertial Systems.}
    \resizebox{\linewidth}{!} {
        \begin{tabular}{lcccccc}
            \toprule
            VIO system & Camera &  Estimation & Pixels used & Output rate & Initialization\\
            \hline
            Basalt & Stereo & Optimization & Feature-based & Frame & Unrestricted\\
            FLVIS & Stereo & Feedback-feedforward loops & Feature-based & Frame/IMU & Unrestricted\\
            Kimera-VIO & Stereo & Optimization & Feature-based & Keyframe & Stationary/Ground-truth\\
            OKVIS & Stereo & Optimization & Feature-based & Frame/IMU & Unrestricted\\
            ORB-SLAM3 & Mono/Stereo & Optimization & Feature-based & Frame & Unrestricted\\
            REBVO & Mono & Optimization & Edge-based & Frame & Unrestricted\\
            ROVIO & Mono/Stereo & Filtering & Direct & Frame & Unrestricted\\
            R-VIO & Mono & Filtering & Feature-based & Frame & Stationary\\
            SVO 2.0 & Mono/Stereo & Optimization  & Semi-direct & Frame & Unrestricted\\
            S-MSCKF & Stereo & Filtering & Feature-based & Frame & Stationary\\
            VINS-Fusion & Mono/Stereo & Optimization & Feature-based & Frame & Unrestricted\\
            \bottomrule
        \end{tabular}
    }
    \label{tab:methods}
\end{table*}
\subsection{Basalt}
Basalt \cite{usenko2020visual} is a visual-inertial system that extracts non-linear factors \cite{mazuran2016nonlinear} from visual-inertial odometry and uses this information to perform mapping. 

The VIO stage is formulated as a fixed-lag smoothing. Initially, in the frontend, KLT-based optical flow \cite{lucas1981iterative} is used to track features in consecutive frames. The backend receives this information and associates features and 3D landmarks to calculate the reprojection error. As for the optimization step, a cost function is defined combining reprojection error terms and IMU error terms, calculated using pre-integration \cite{forster2015imu}. Finally, a marginalization of old states is applied in order to keep the problem computationally tractable. Non-linear factors are extracted after marginalizing a keyframe. These factors contain visual-inertial information about the camera motion. 

In the mapping stage, a bundle adjustment is performed to refine camera poses of keyframes and landmarks positions. In order to accomplish this, new features that are more suitable for loop closure detection are extracted and matched. Then, the reprojection error is calculated from this new keypoints. Non-linear factors extracted in the VIO stage and the reprojection error are combined in a new cost function, which is minimized. The result of this optimization is a globally consistent map.
\subsection{FLVIS}
FLVIS (Feedforward-feedback Loop-based Visual Inertial System) \cite{chen2020flvis} is a visual-inertial pose estimation system that takes ideas from control theory and signal processing and applies them to the VIO problem.

The IMU measurements are used in the frontend of the system. On the one hand, a frame-to-frame tracking is performed by extracting features and tracking them with KLT. After doing this, the IMU information is incorporated (feedforward loop) to compensate roll and pitch before an in-frame optimization is executed. Only the pose is optimized, leaving the landmarks fixed. From the result of this optimization, the IMU bias is estimated and updated (feedback loop). The positions of landmarks are updated through an Infinite Impulse Response (IIR) filter.

The backend performs an optimization on a sliding window of keyframes that tries to minimize only the reprojection error of the landmarks and passes this correction to the frontend. The system also has a loop closure module implemented as a pose graph optimization.
\subsection{Kimera-VIO}
The Kimera library \cite{rosinol2020kimera} is composed of different modules. Kimera-VIO is a VIO module based on the work of \citet{forster2017onmanifold}. The frontend uses the Shi-Tomasi detector \cite{shi1994good} and KLT \cite{lucas1981iterative}. Feature detection, stereo matching and geometric verification are implemented only for keyframes. KLT-based tracking is done for all intermediate frames. The backend uses a GTSAM-based \cite{dellaert2012factor} fixed-lag smoother (it can be configured to be full smoothing) in which the pre-integrated IMU information between keyframes and visual measurements are incorporated. Degenerate points (behind the camera or without enough parallax) and outliers are removed before eliminating 3D points from the VIO state \cite{carlone2014eliminatingci}.

The Kimera-RPGO module is in charge of detecting loop closures and computing keyframe poses in a globally consistent way. RPGO (Robust Pose Graph Optimization) uses a PCM-based outliers rejection method \cite{mangelson2018pairwise}. Kimera-Mesher, that computes a fast 3D mesh to support obstacle avoidance, and Kimera-Semantics, that builds a more accurate global 3D mesh semantically annotated, are the other modules that compose the library.
\subsection{OKVIS}
Open Keyframe-based Visual-Inertial SLAM (OKVIS) \cite{leutenegger2015keyframe} is a visual-inertial algorithm based on optimization. The method proposes a cost function to be optimized which is composed of the reprojection error of the landmarks together with an IMU error term. A combination of multi-scale SSE-optimized Harris corner detector \cite{harris88acombined} and BRISK descriptor \cite{leutenegger2011brisk} is implemented in the frontend. A keyframe-based optimization window is used instead of processing poses from consecutive frames over time. Since keyframes may be spaced arbitrarily far in time, this approach can be beneficial when the camera exhibits little or no motion. As a last step, marginalization is applied to remove old states and allow OKVIS to run in real time.
\subsection{ORB-SLAM3}
ORB-SLAM3 \cite{campos2020orbslam3} is a system based on ORB-SLAM2 \cite{mur2017orb} and ORB-SLAM-VI \cite{mur2017visual}. It implements an initialization based on Maximum a Posteriori (MAP) estimations, which is faster than the initialization of ORB-SLAM-VI. The pipeline is divided into three main threads. The tracking thread determines the frame pose in the current map minimizing the reprojection errors between the last two frames, using ORB features, and decides if a frame becomes a keyframe. The local mapping thread refines the map locally by making a visual-inertial bundle adjustment in a window of keyframes. The loop and map merging thread recognizes previously visited places and performs a loop closure.
The backend performs MAP estimation, minimizing visual residuals (reprojection errors) and inertial residuals based on the pre-integration of IMU measurements between frames \cite{forster2017onmanifold}.
\subsection{REBVO}
REBVO (Realtime Edge-Based Visual Odometry) \cite{tarrio2015realtime} is a midpoint between feature-based and direct methods. First, the edges are extracted from the input image. Unlike classic features, the edges provide structural information of the observed scene. From this extraction an edge-map is obtained, which is made up of keylines.

In a second stage a SE(3) transformation is estimated between the edge-map of the previous frame and the current frame. A cost function is defined to minimize for each edge the distance to the nearest edge after reprojection.
Finally, a matching is made between the keylines of the current edge-map and the keylines of the previous edge-map. Extended Kalman Filter (EKF) is used to estimate and update the depth of each keyline in the new edge-map. 

As for the IMU information, it is used to estimate a rotation with which the edge-map of the previous frame is pre-rotated, as a previous step to the estimation of the SE(3) transformation. Additionally, a prior can be added to the cost function with IMU information, similar to SVO 2.0.
\subsection{ROVIO}
Robust Visual-Inertial Odometry (ROVIO) \cite{bloesch2015robust} is a filter-based estimator.  The traditional approach when EKF is implemented to solve the SLAM problem uses IMU measurements to propagate the state of the filter, while the visual information is taken into account during the update step. ROVIO includes visual features into the state vector of the EKF. Initially, FAST \cite{rosten2006machine} corner features are extracted, and multi-level patches are obtained from these features. These patches allow the filter to work with the pixel intensity error in the update step (direct method). Another important property of ROVIO is that it implements a robocentric approach: the 3D position of the features is always estimated with respect to the current camera pose. According to the authors, this approach improves the consistency of the filter, since it avoids problems with unobservable states.
\subsection{R-VIO}
R-VIO (Robocentric Visual-Inertial Odometry) \cite{huai2018robocentric} is a filter-based VIO algorithm that implements a robocentric approach. The motion information of the starting frame is part of the state vector. Unlike ROVIO, R-VIO does not maintain the observed features in the state vector. Instead, it only keeps a small amount of relative poses. The frontend uses the Shi-Tomasi corner detector \cite{shi1994good} to find features and KLT to track them \cite{lucas1981iterative}. After studying the observability properties of the proposed approach, the authors conclude that it does not present the observability issue that is characteristic of other filter-based methods \cite{li2013high, huang2010observability}.
\subsection{SVO 2.0}
SVO 2.0 \cite{forster2017svo} is an extension of the original SVO (Semi-direct Visual Odometry) implementation presented by \citet{forster2014svo}. It is a VIO system based on optimization that incorporates information from other sensors as motion prior. In general it runs faster than similar state-of-the-art systems because it was designed to be computationally efficient. The method decouples tracking from mapping by using two parallel threads. 

The motion estimation thread estimates the relative transformation between the current image and the last frame by minimizing the photometric error. This optimization is performed only for those pixels in the image corresponding to corners and edgelets. These feature pixels are extracted in the mapping thread for keyframes only. Their depth is obtained using a recursive Bayes filter. Once the depth is calculated, these points are added to the map and can be used to calculate the pose in the motion estimation thread. As for the IMU measurements, they are added as motion prior during the minimization of the photometric error.
\subsection{S-MSCKF}
Stereo Multi-State Constraint Kalman Filter (S-MSCKF) \cite{sun2018robust} is a stereo VIO algorithm based on the EKF. This method takes the work of \citet{mourikis2007multistate} as a starting point. In the frontend, KLT-based optical flow \cite{lucas1981iterative} is used to track features temporally and to match stereo features. While KLT-based methods are not as accurate for temporal tracking as descriptor-based methods \cite{paul2017comparative}, the authors find that they require much less CPU resources which is what they are more interested in because they have to handle stereo images. The state vector contains the IMU state and a fixed number of camera states but not the observed features. Once the number of camera states reaches a threshold, some of them are marginalized. Finally, to improve the consistency of the filter, Observability Constrained EKF \cite{huang2010observability} is implemented.
\subsection{VINS-Fusion}
VINS-Fusion \cite{qin2018online, qin2019general} is an optimization-based state estimator. It is an extension of VINS-Mono \cite{qin2018vins} that supports multiple sensors and different configurations (stereo cameras, mono camera with IMU and stereo cameras with IMU).

During the initialization process, VINS-Fusion combines visual information with pre-integration \cite{forster2015imu} of the IMU using a loosely alignment. First, a visual-only SfM (Structure from Motion) is used to calculate up-to-scale camera poses and feature positions. This is done in a sliding window. In a second step, information from the IMU is incorporated to obtain an initial calibration of the gyro bias, and initial values for the velocities of each frame of the window, gravity vector and scale. 

Unlike OKVIS, VINS-Fusion pre-integrates the IMU measurements to build the IMU error term of the cost function. The system also has a relocalization module to eliminate drift. It finds correspondences between features, which are incorporated into the cost function to be optimized.

As a clarification with respect to the output rate of VINS-Fusion in Table~\ref{tab:methods}, if multiple threads are used, for every two input frames, only one pose is estimated.

\section{Experimental Evaluation and Discussion}
\label{sec:experiments}
In this section we show the evaluation results of the selected VIO systems (described in Section~\ref{sec:method}) in a soybean field. In particular, we used our recently released Rosario Dataset~\cite{pire2019rosario}, recorded with the sensors of our weed removal robot while it was manually guided with a remote controller.

\subsection{The Rosario Dataset: issues found and improvements made}
The Rosario Dataset~\cite{pire2019rosario} is a collection of sensor data recorded in agricultural scenes. It is composed of six sequences captured in a soybean field by the weed removal robot built by the CIFASIS (CONICET-UNR). The dataset contains \SI{15}{\Hz} stereo images, \SI{142}{\Hz} IMU measurements, wheel odometry and RTK-GNSS. The RTK-GNSS data are used as positional ground-truth. The soybean environment recorded in the Rosario Dataset presents several challenges for localization and mapping systems. It is characterized by its highly repetitive visual appearance, its irregular terrains and a variety of lighting conditions.

During the experiments, we detected two different types of limitations regarding the collected IMU measurements. They could be separated into magnitude and timestamp problems. The first ones are due to how the IMU was configured for the data recording. In order to get more resolution, the measurement range for accelerations was limited to approximately $\pm \SI{19.6}{\metre\per\second\squared}$. Unfortunately, the shaking movement caused by the uneven soil made the accelerometer reach its positive saturation value along the gravity parallel axis during some instants. This condition makes some of the least robust studied VIO systems fail with this kind of short-lived perturbations. 

Regarding timestamps, the limitation is a consequence of the software synchronization explained in \citet{pire2019rosario}, that was made at the level of user application in the operating system. Because of the high output data rate of the IMU, some intermediate buffering was needed while recording, causing time gaps without IMU measurements and time gaps with abnormally concentrated IMU measurements. Notice that no measurements were lost in the process. Considering this, we fixed it by taking advantage of the fact that the output data rate of the IMU was known and stable enough because of its internal oscillator that keeps the sample period fixed. This allowed us to temporally redistribute the data with its correct frequency. 

There is an additional thing that was fixed previously to evaluate the VIO systems. We found a bug in the recording software related to a wrong indexing of a circular buffer, that rendered useless the last parts of the sequences 01 and 06. Having realized that, we trimmed them eliminating the inaccurate parts.

These fixes were uploaded to the Rosario Dataset's website \footnote{\href{https://www.cifasis-conicet.gov.ar/robot/}{\nolinkurl{https://www.cifasis-conicet.gov.ar/robot/}}} so that the community may benefit from them.
%

\subsection{Evaluation of VIO systems in the Rosario Dataset}
An evaluation of the systems presented in section~\ref{sec:method} was performed on the Rosario Dataset. All the experiments have been run using an Intel\textsuperscript{\textregistered} Core i9-9900K CPU @ \SI{3.60}{\giga\hertz}, with \SI{64}{\giga\byte} of RAM memory. 


In order to facilitate both the installation of the systems and the reproducibility of the results, we ran the systems inside Docker containers \cite{merkel2014docker}. Docker also helps to encapsulate the dependencies of each system, keeping us away from the problem of having to install different versions of a library in the same computer. In addition, we used ROS system implementations to simulate the online processing on the robot, which is imperative since we will use the estimated pose to feed the robot control which needs to work in real time. ROS also gave us a fair and straightforward way to unify the comparison of processing time and accuracy among the different systems. It is important to note that while most of the systems provide a ROS interface or wrapper to run them, we adapted and fixed those which do not support ROS or do not do it in a suitable way. This is the case of Basalt and REBVO for which we adjusted and fixed the provided ROS wrappers code; and ORB-SLAM3 for which the ROS wrapper does not publish any localization information until the end of the run. As we have mentioned, our final goal is to control the robot in real time so the estimated pose should be accessible while the VIO system is running. Considering this, we modified the ORB-SLAM3 ROS wrapper to publish the estimated pose as soon as a frame is properly tracked.

With respect to the settings of the methods, all of them are available in the repository previously referenced in Section~\ref{sec:introduction}. Whenever possible, we chose a Stereo-Inertial configuration. Notice that REBVO and R-VIO only support Monocular-Inertial mode (see Table~\ref{tab:methods}). The configuration parameters of each system were tuned, up to a feasible point, to improve the accuracy of the pose estimation. The IMU noise parameters were one of the specific adjustments we did. Some systems are more sensitive to them than others, so in those cases we increased the values by one or two orders of magnitude.



Since Kimera-VIO, S-MSCKF and R-VIO systems require the robot to start stationary for initialization, we would not be able to evaluate them properly in sequences 01, 02, 05 and 06, which start with the robot under motion. To have an evaluation of each system in the largest number of sequences, we take advantage of the fact that sequences 01 and 05 have a part where the robot stops for a couple of seconds, to generate two new sequences from those stopping moments. Unfortunately, this procedure could not be repeated for the sequences 02 and 06, since either the robot does not stop in any time, or the resulting trimmed sequence would be too short.

Given that the Rosario Dataset has positional ground-truth (no orientation is provided), we adopt the Absolute Trajectory Error (ATE) \cite{sturm2012benchmark} as metric to evaluate systems accuracy. Furthermore, since there are no loops in the dataset, a VI-SLAM system such as ORB-SLAM3, i.e. one that reuses the map to improve localization and reduce drift, can be fairly compared to VIO systems. We ran the systems multiple times to check the consistency and repeatability of the results. Then, for each case the errors were calculated using the benchmark suite tool evo \cite{grupp2017evo} and the minimum ones were chosen to be reported. Table~\ref{tab:aperpe} shows their means and standard deviations. ATE has been computed after the estimated trajectories were aligned with the ground-truth using Umeyama alignment \cite{umeyama1991least}.


%
\begin{landscape}
    \begin{table}
        \centering
        \caption{Mean and standard deviation (between parentheses) of the Absolute Trajectory Error (ATE) [\si{\metre}] for the VIO systems in the sequences of the Rosario Dataset. Minimum errors per sequence are depicted in bold.}
        \begin{tabular}{lcccccccc}
            \toprule
            \multirow{2}{*}{VIO system} & \multicolumn{8}{c}{Sequence} \\
            \cmidrule{2-9}
                            & 01                    & 02                    & 03                & 04                    & 05                    & 06                    & Trimmed 01        & Trimmed 05            \\
            \midrule
            Basalt          & 14.68 (6.04)          & 10.01 (3.79)          & 7.87 (4.41)       & 8.20 (4.64)           & 10.42 (5.45)          & 22.49 (11.06)         & 12.2 (3.97)       & 8.51 (4.80)           \\
            FLVIS           & 20.4 (11.3)           & 7.42 (3.88)           & 5.77 (2.83)       & 5.77 (2.77)           & 17.1 (11.4)           & 25.7 (14.4)           & 17.3 (8.09)       & 7.38 (3.76)           \\
            Kimera-VIO      & -                     & 7.54 (3.22)           & -                 & 6.77 (3.63)           & -                     & -                     & 11.1 (3.77)       & 7.47 (3.82)           \\
            OKVIS           & 6.54 (3.24)           & 7.75 (4.69)           & 7.70 (4.46)       & 7.41 (4.26)           & 7.80 (4.53)           & 15.9 (9.21)           & 6.55 (3.22)       & 8.64 (5.02)           \\
            ORB-SLAM3       & {\bf 1.10} (1.23)     & 1.79 (1.66)           & 2.20 (2.67)       & 2.61 (2.77)           & 2.05 (2.73)           & {\bf 3.06} (2.06)     & 2.24 (1.93)       & 1.06 (0.64)           \\
            REBVO           & -                     & 9.05 (5.59)           & 8.70 (5.97)       & 5.55 (4.63)           & -                     & -                     & -                 & 8.00 (4.39)           \\
            ROVIO           & 5.48 (2.68)           & 7.44 (3.99)           & 5.32 (2.96)       & 4.98 (2.78)           & 5.99 (3.61)           & 12.1 (6.29)           & 6.06 (3.31)       & 5.98 (3.52)           \\
            R-VIO           & -                     & 2.68 (1.38)           & -                 & 2.21 (1.11)           & -                     & -                     & 5.32 (3.01)       & -                     \\
            SVO-2.0         & 12.3 (5.20)           & 7.78 (4.89)           & 6.63 (3.56)       & 5.94 (3.40)           & 11.27 (6.88)          & 17.2 (8.73)           & 12.2 (4.94)       & 7.32 (3.58)           \\
            S-MSCKF         & 3.11 (2.17)           & {\bf 1.44} (1.19)     & {\bf 0.50} (0.22) & {\bf 0.50} (0.24)     & {\bf 1.55} (0.88)     & 4.24 (2.55)           & {\bf 1.97} (1.49) & {\bf 0.61} (0.23)     \\
            VINS-Fusion     & 9.68 (6.09)           & 13.3 (8.95)           & 7.42 (6.29)       & 3.70 (2.09)           & 24.1 (18.2)           & 9.35 (7.13)           & 6.39 (2.81)       & 3.48 (2.23)           \\
            \bottomrule
        \end{tabular}
        \label{tab:aperpe}
    \end{table}
    
    \begin{table}
        \centering
        \caption{Mean and standard deviation (between parentheses) of the Absolute Trajectory Error (ATE) [\si{\metre}] for ORB-SLAM3 and S-MSCKF in the sequences of the Rosario Dataset after ORB-SLAM3 initialization. Minimum errors are depicted in bold.}
        \begin{tabular}{lcccccccc}
            \toprule
            \multirow{2}{*}{VIO system} & \multicolumn{8}{c}{Sequence} \\
            \cmidrule{2-9}
                            & 01                    & 02                    & 03                & 04                    & 05                    & 06                    & Trimmed 01        & Trimmed 05            \\
            \midrule
            ORB-SLAM3       & {\bf 0.86} (0.44)     & 1.43 (0.81)           & 0.99 (0.63)       & 1.10 (0.67)           & {\bf 1.07} (0.56)     & {\bf 2.74} (1.65)     & {\bf 0.91} (0.43) & 1.06 (0.64)           \\
            S-MSCKF         & 3.13 (2.16)           & {\bf 1.21} (1.05)     & {\bf 0.40} (0.17) & {\bf 0.41} (0.20)     & 1.41 (0.79)           & 3.85 (2.88)           & 1.34 (0.95)       & {\bf 0.61} (0.23)     \\
            \bottomrule
        \end{tabular}
        \label{tab:orbaperpe}
    \end{table}
\end{landscape}

\begin{figure}[!ht]
  \centering
  \subfloat[\label{fail_rebvo}]{\includegraphics[width=1\columnwidth]{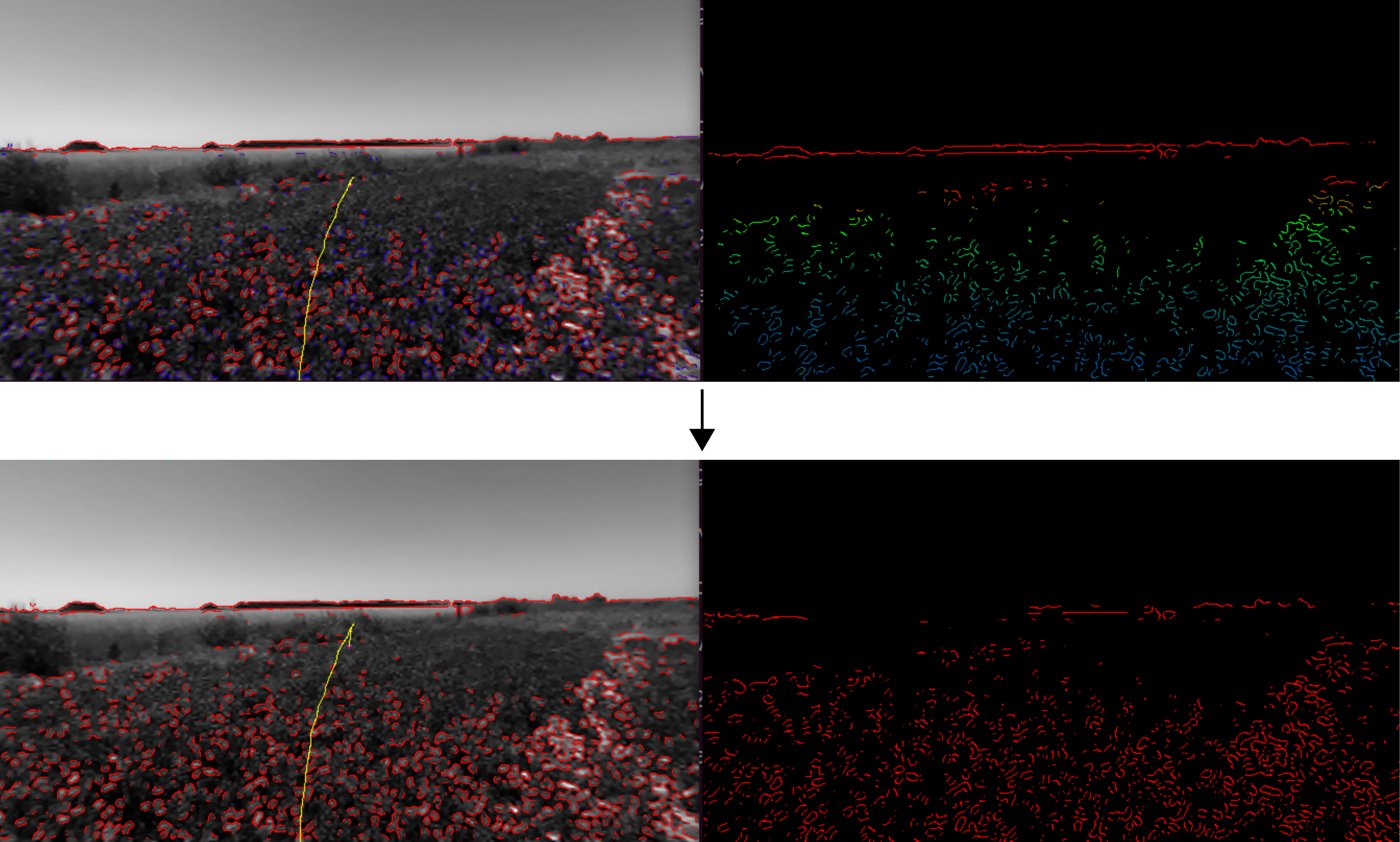}\label{fail_rebvo_int}}\\
  \subfloat[\label{fail_kimera}]{\includegraphics[width=0.55\columnwidth]{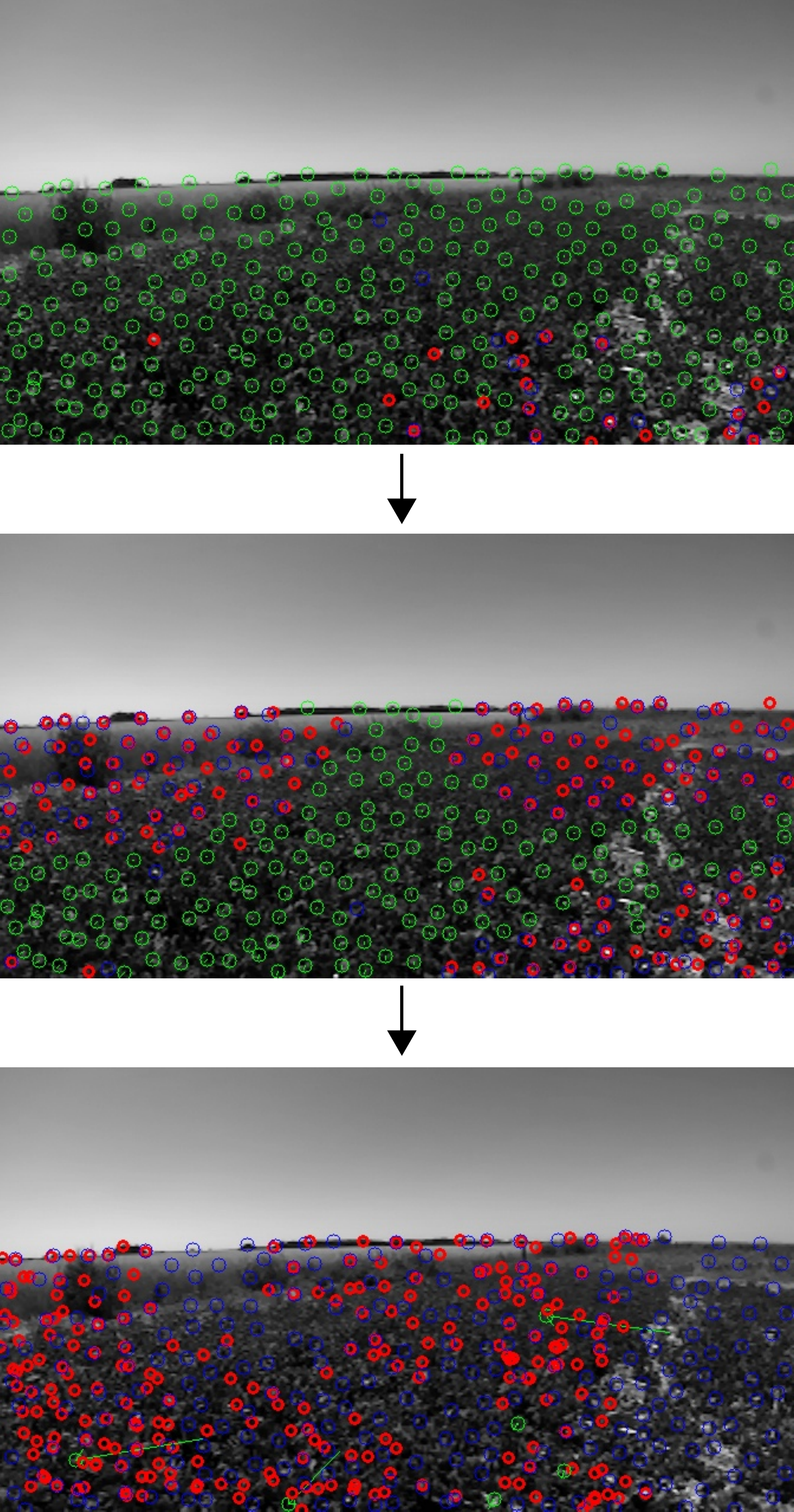}\label{fail_kimera_int}}\\
  \subfloat[\label{fail_okvis}]{\includegraphics[width=0.55\columnwidth]{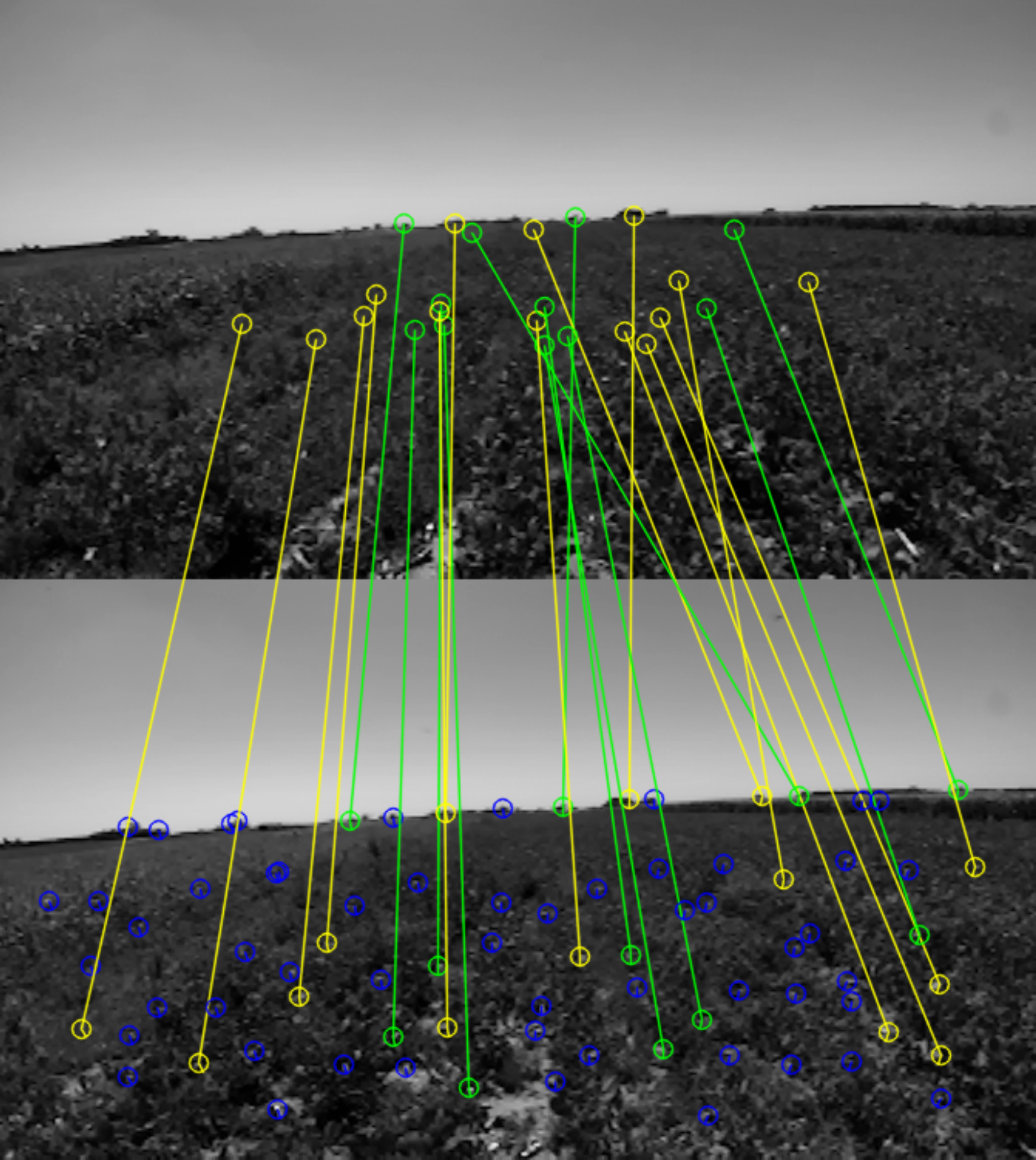}\label{fail_okvis_int}}\\
  \caption{Examples of tracking failures of the assessed VIO systems (using the provided visualization interface). \protect\subref{fail_rebvo_int} REBVO, \protect\subref{fail_kimera_int} Kimera-VIO and \protect\subref{fail_okvis_int} OKVIS.}
  \label{fig:fails}
\end{figure}

For the sake of completeness, we include the results of all the systems in the eight sequences, even when they have special requirements to guarantee a good initialization as in the cases of Kimera-VIO, R-VIO and S-MSCKF.
Kimera-VIO is supposed to offer three ways to auto initialize but at the moment of testing, only two were implemented. These ones assume the robot being stationary in the beginning or the availability of ground-truth data (including pose and linear velocity). None of those conditions were fulfilled.
For their part, the authors of R-VIO clarify that their implementation is only to validate the approach proposed in \citet{huai2018robocentric}, so the initialization was not a main concern and it is rather simple. It also assumes a stationary state in the beginning.
Being said this, the failures of Kimera-VIO and R-VIO in three out of the four sequences that start under motion were expected and confirmed that their initialization procedures are not robust enough.

As opposed to expected, the cases of S-MSCKF and REBVO should be mentioned. Despite of its stationary beginning requirement, S-MSCKF was able to initialize properly under motion obtaining impressive results. This is explained because S-MSCKF uses an Observability Constrained EKF (OC-EKF) \cite{hesch2012ocekf} to maintain the consistency of the filter which makes it less dependent on an accurate initialization \cite{sun2018robust}. On the other hand, REBVO, which do not require particular conditions for initialization, failed to initialize in sequence 01 and 05. In sequence 01, REBVO fails when the robot starts both in motion and in a stationary state. With respect to sequence 05, REBVO is affected at the beginning of the trajectory by IMU saturation, which is a consequence of the uneven terrain. When REBVO is executed from a time instant after saturation, it is able to initialize properly and complete sequence 05.

Figure~\ref{fig:fails} shows the tracking failures hit by the assessed VIO systems. REBVO and Kimera-VIO fail in the same part of the sequence 03, in which the robot stops for a moment. In Figure~\ref{fail_rebvo}, it can be seen that when it starts moving again, all the previously detected edges become red indicating that REBVO lost the scene depth. From this moment onward, even though the system kept working, the quality of its estimation decreased considerably. Figure~\ref{fail_kimera} shows how features are lost or wrongly detected by Kimera-VIO also when the robot restores its movement. Finally, in Figure~\ref{fail_okvis} wrong matches detected by OKVIS between landmarks (projected onto last keyframe in the upper image) and current features are presented. This corresponds to the first part of the sequence 05 and is related to a wrong initialization caused by the robot's movement from the very beginning of the sequence. Notice that this situation does not always occur.

As can be observed in Table~\ref{tab:aperpe}, S-MSCKF and ORB-SLAM3 outperform notoriously the rest of the systems. With respect to robustness, ROVIO, Basalt and FLVIS are also remarkable. The experiments were repeated several times and only these five systems were always able to complete the sequences. VINS-Fusion occasionally fails in sequence 01 and the same happens with OKVIS and SVO 2.0 in sequence 05. The rest of the methods systematically fail in at least one sequence. 

\begin{figure*}[!htp]
  \centering
  \subfloat[Sequence 01]{\includegraphics[width=0.42\textwidth]{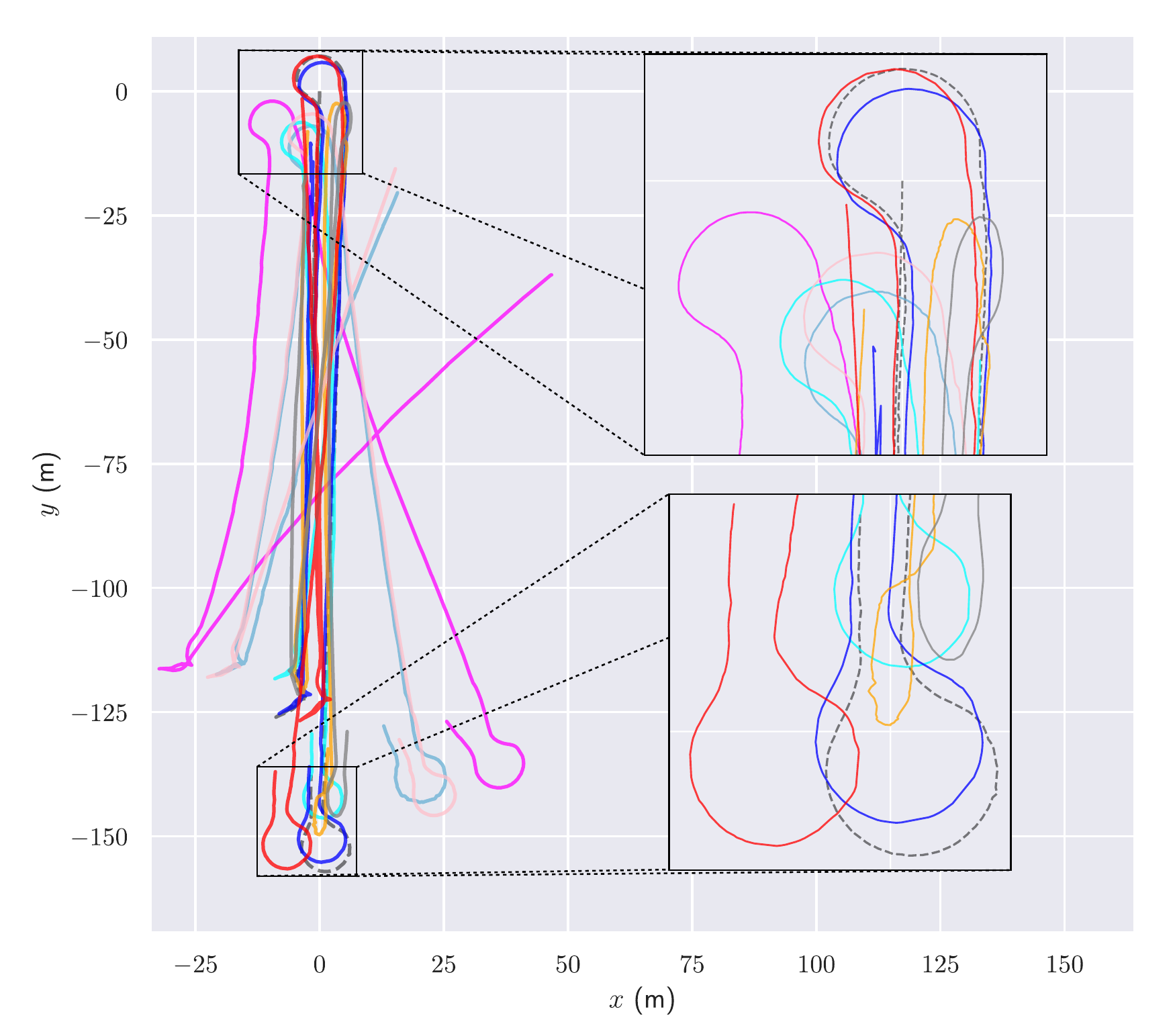}}
  \hspace{1cm}
  \subfloat[Sequence 02\label{plot_legends}]{\includegraphics[width=0.42\textwidth]{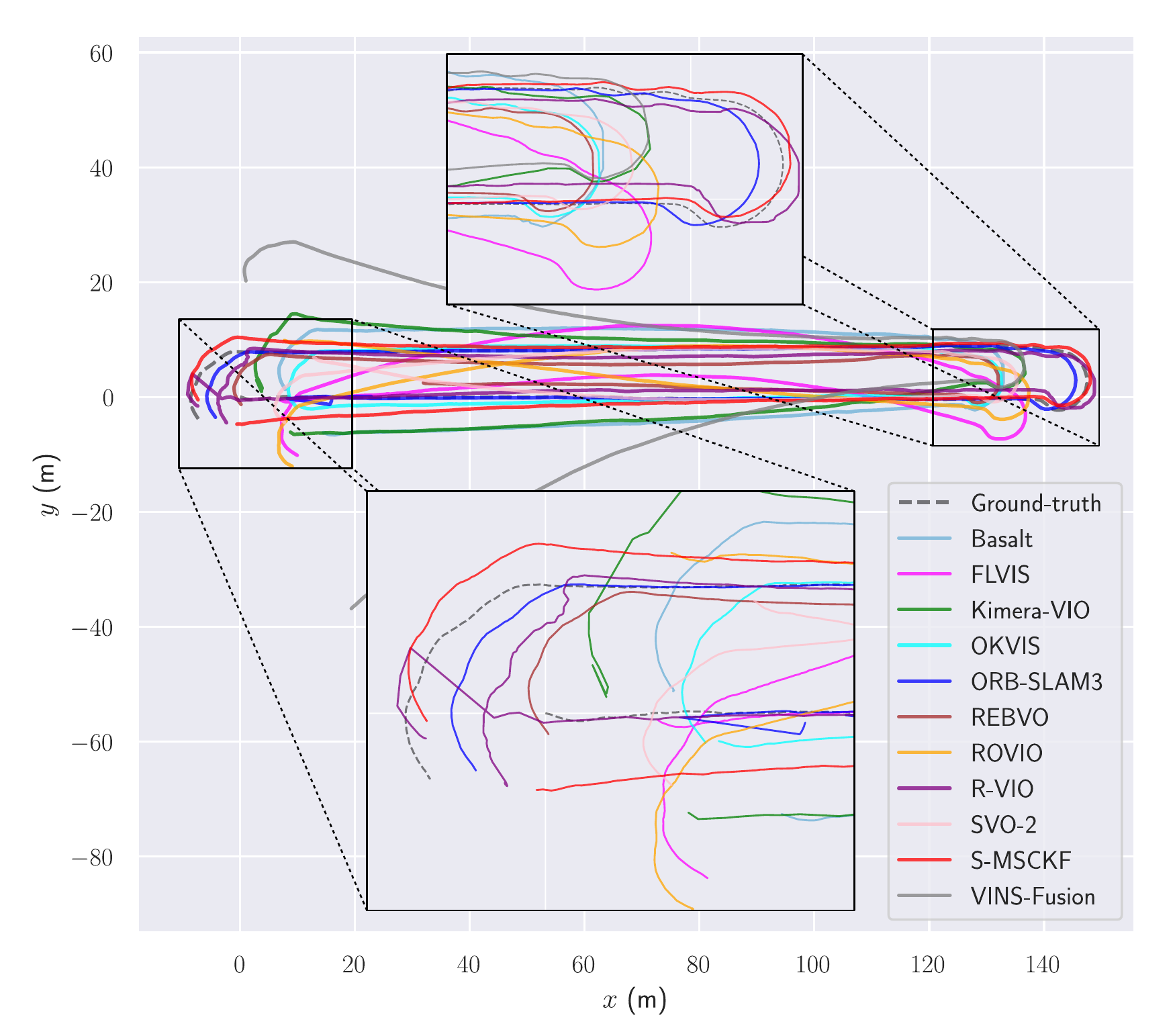}\label{plot_legends_int}}\\
  \subfloat[Sequence 03]{\includegraphics[width=0.42\textwidth]{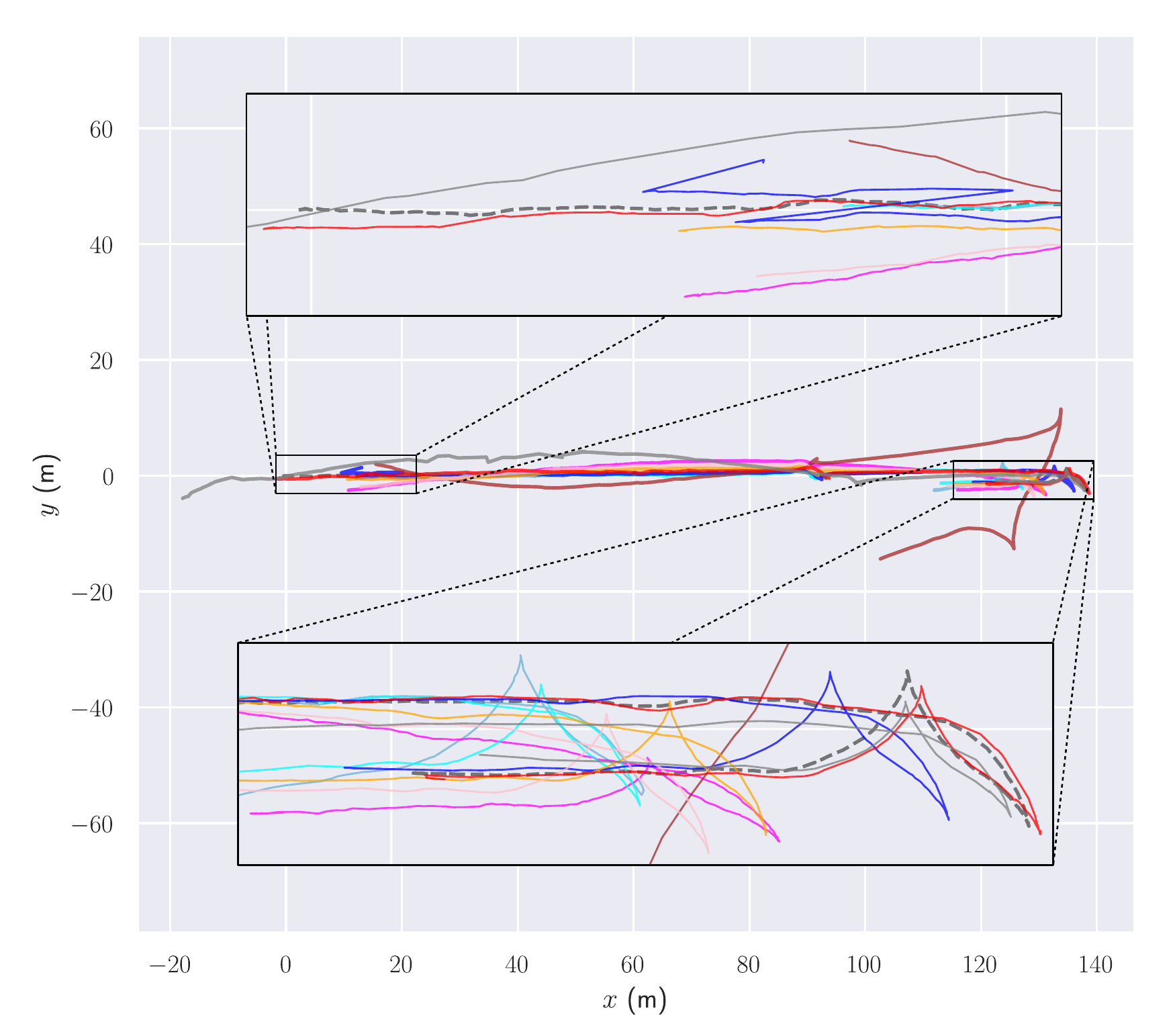}}
  \hspace{1cm}
  \subfloat[Sequence 04]{\includegraphics[width=0.42\textwidth]{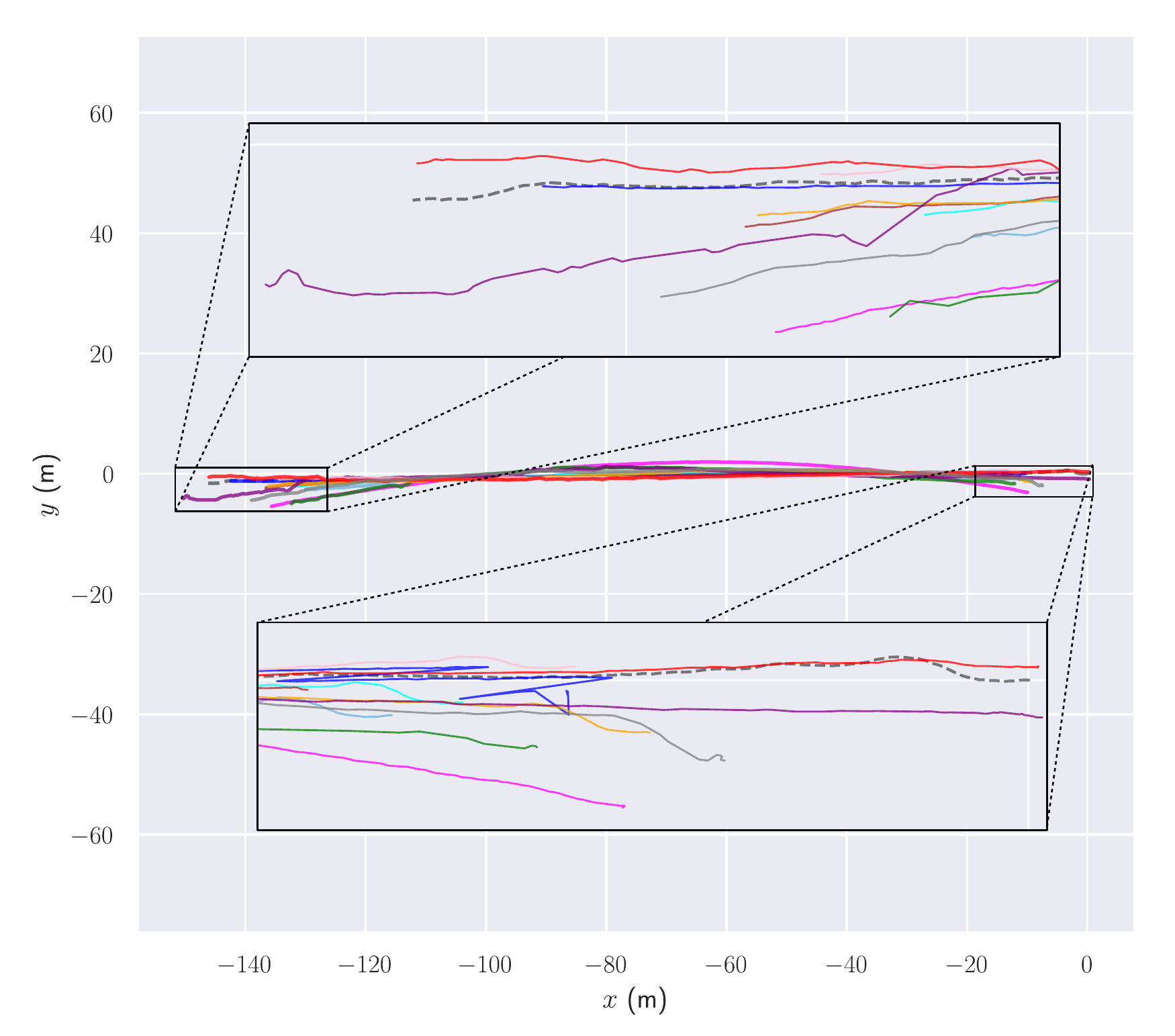}}\\
  \subfloat[Sequence 05]{\includegraphics[width=0.42\textwidth]{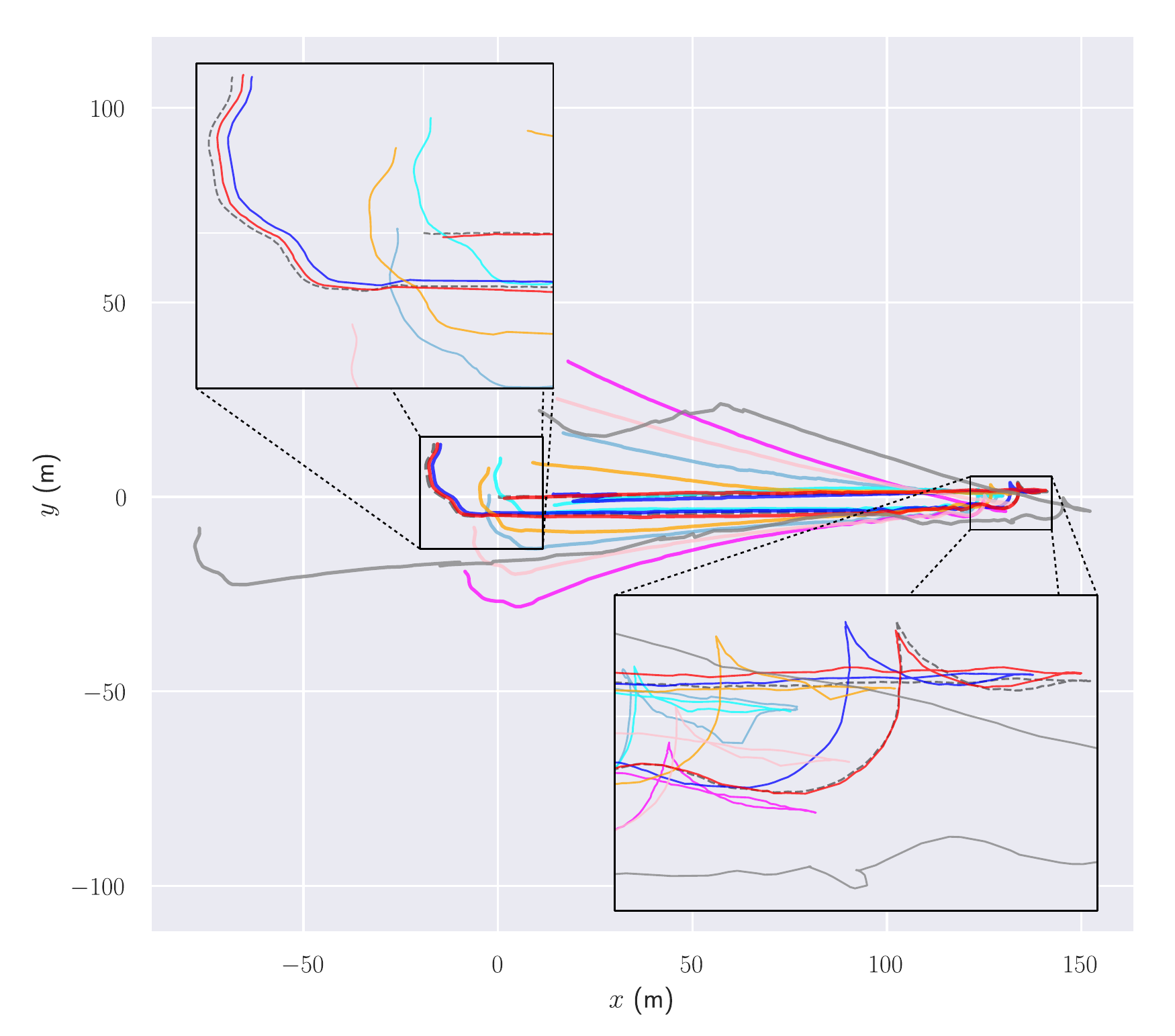}}
  \hspace{1cm}
  \subfloat[Sequence 06]{\includegraphics[width=0.42\textwidth]{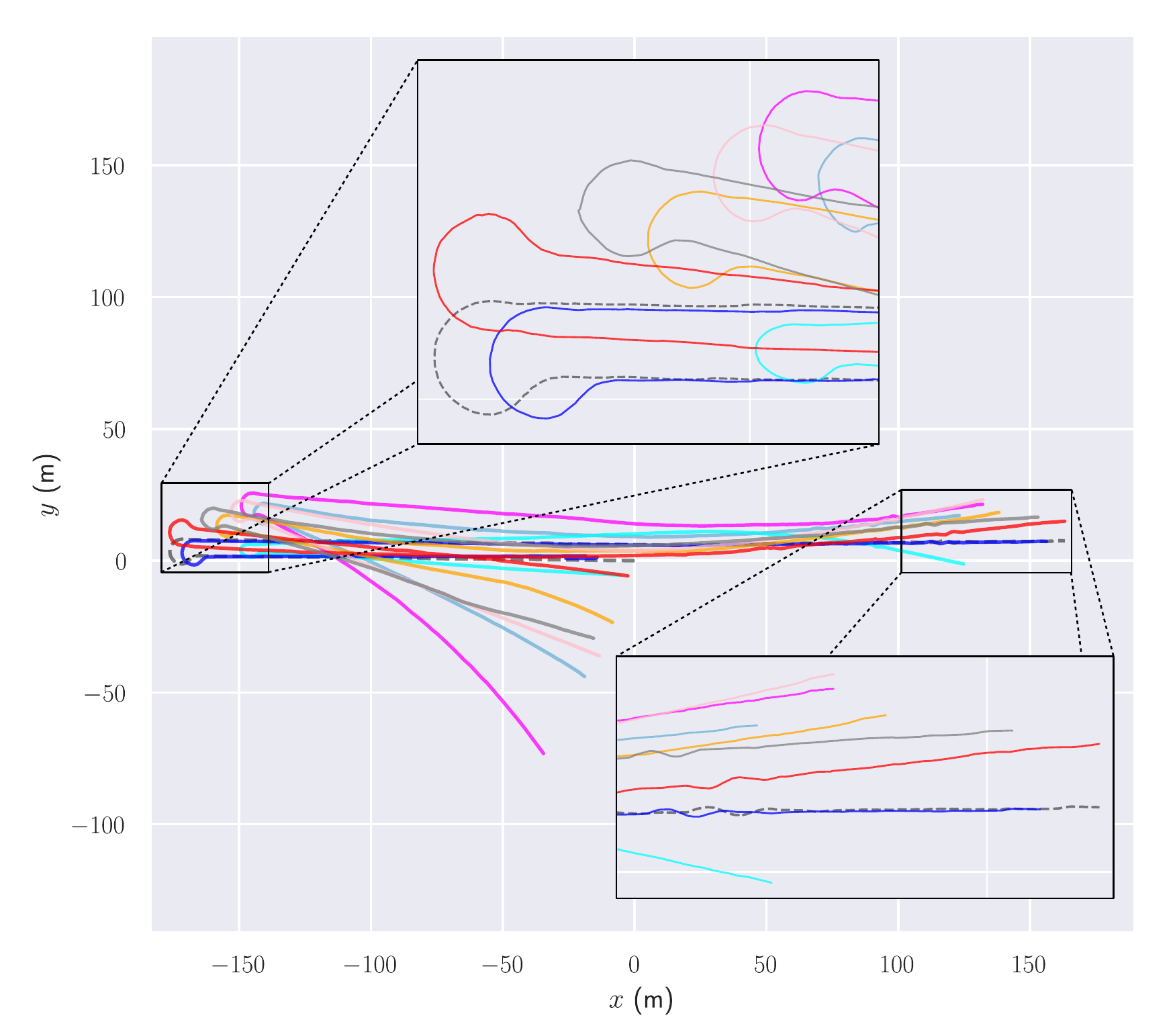}}
  \caption{Trajectories generated by the VIO systems on the Rosario Dataset sequences. They are globally aligned. Plots keep the aspect ratio and legend labels shown in \protect\subref{plot_legends_int} are valid for all of them.}
  \label{fig:trajectories}
\end{figure*}

\begin{figure*}[!htp]
  \centering
  \subfloat[Trimmed sequence 01]{\includegraphics[width=0.42\textwidth]{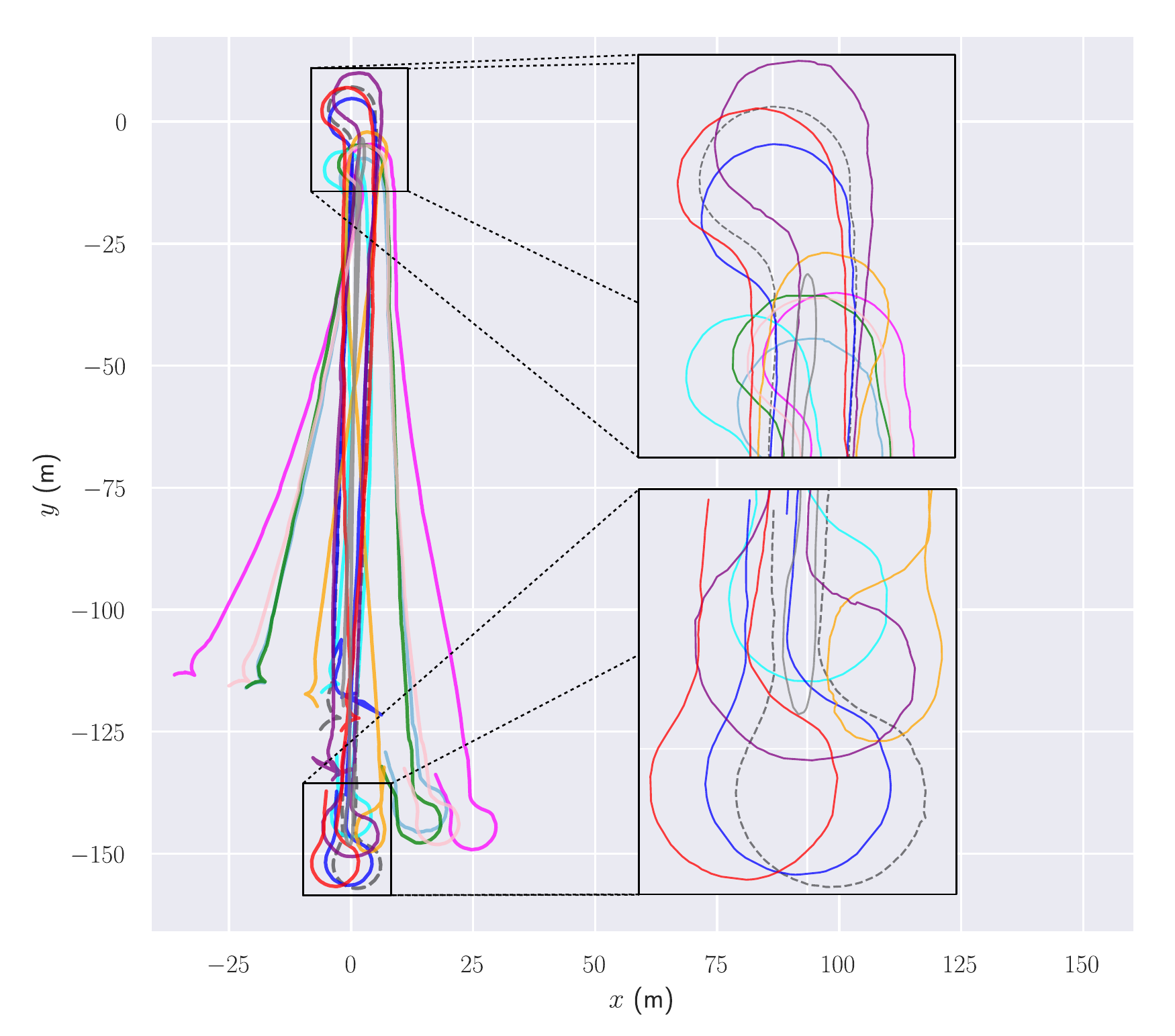}}
  \hspace{1cm}
  \subfloat[Trimmed sequence 05]{\includegraphics[width=0.42\textwidth]{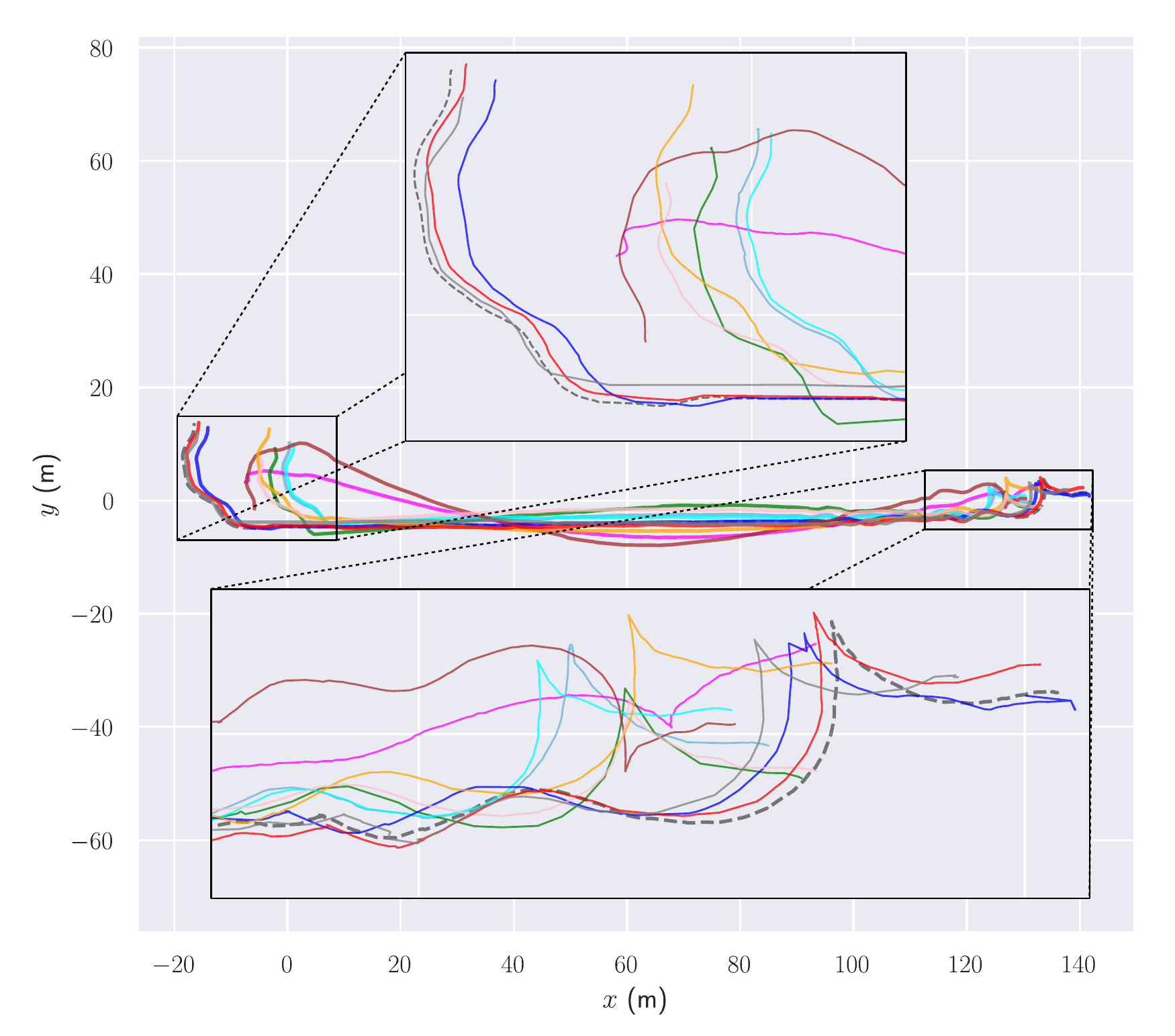}}
  \caption{Trajectories generated by the VIO systems on the trimmed sequences 01 and 05 (from stationary states). They are globally aligned. Plots keep the aspect ratio and legend labels shown in Figure~\ref{plot_legends} are valid.}
  \label{fig:trajectories_cut}
\end{figure*}

The estimated trajectories can be seen in Figure~\ref{fig:trajectories} and Figure~\ref{fig:trajectories_cut}.
We noticed that most of the ORB-SLAM3 estimation errors occur during the initialization stage. Therefore, to better understand how much this was affecting, we discarded the estimated poses during initialization and recalculated ATE. Table~\ref{tab:orbaperpe} shows these results, including the corresponding ones for S-MSCKF to compare.


It should be noted, as previously mentioned, that the trajectories estimated by ORB-SLAM3 are not the ones obtained once all the data has been processed, as in the original implementation, but the ones produced on the fly. Table~\ref{tab:orbaperpe} summarizes the most important aspect of the evaluation in terms of errors for the application that we are finally interested in, which is autonomous navigation through field crops. The initialization could be ignored since the robot can stay still until that stage is finished and, taking that into account, Table~\ref{tab:orbaperpe} shows the systems that performed best. However, another important aspect of the evaluation also related to the aforementioned task, is the analysis about computational load which is tackled below.

\begin{figure*}[!btp]
    \centering
    \includegraphics[width=\linewidth]{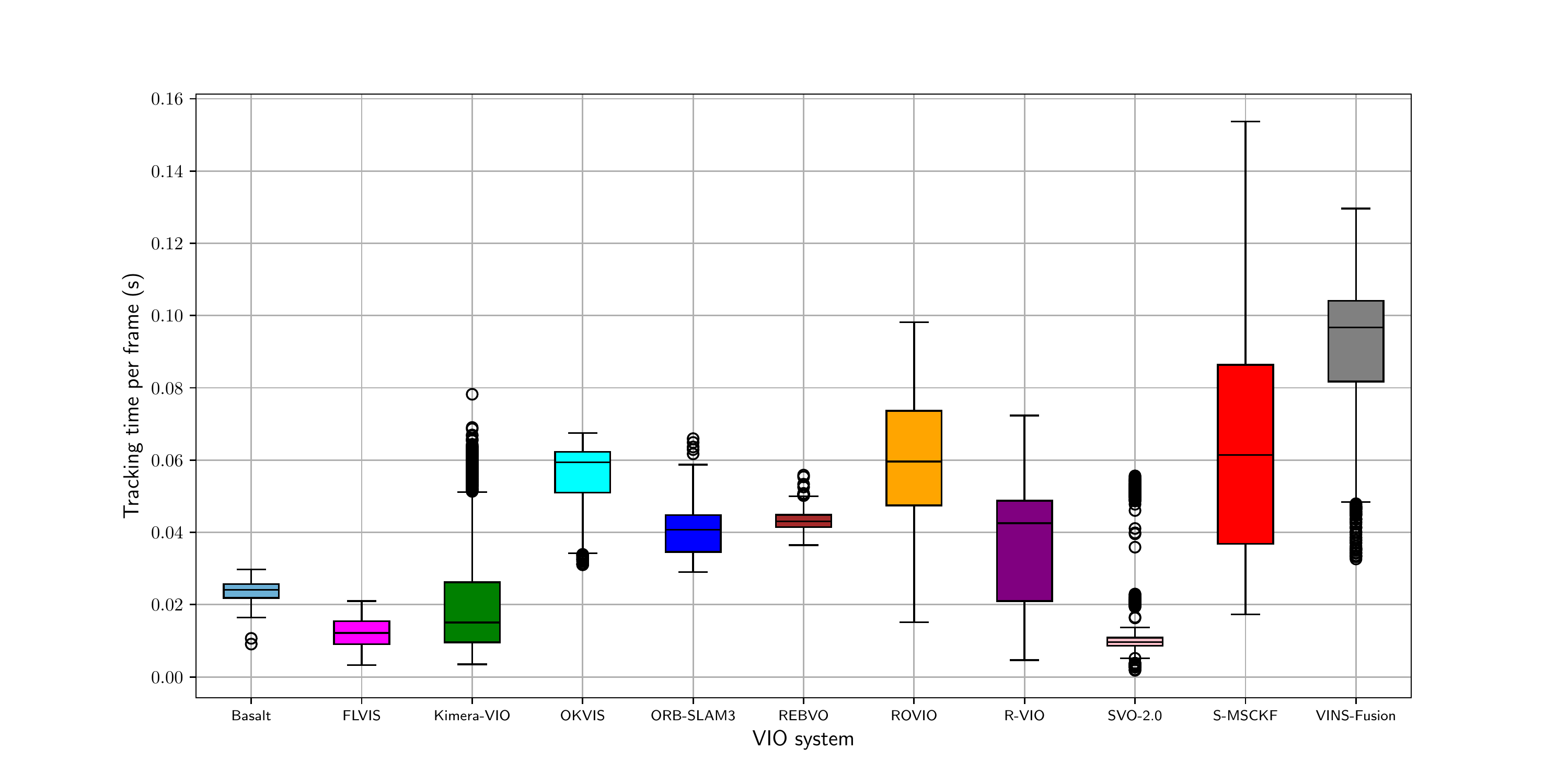}
    \caption{Tracking time comparison.}
    \label{fig:times}
\end{figure*}

Since we assessed several VIO systems conceptually different, to gain insight into how much computation each system requires we measure the tracking time, i.e. the time between the arrival of a stereo image and the first published pose that was obtained taking into account that information (and all the IMU messages that correspond). To do this, the initialization procedures were avoided and we assured that no stereo frame was lost, so that no errors in the association between input frames and output poses were made. The case of Kimera-VIO deserves a special clarification because it publishes a pose after each keyframe is processed instead of publishing it after handling each frame, performing an optimization within the backend that considers the information from multiple previous frames. To make the comparison fairer, for Kimera-VIO we decided to measure the time between the arrival of a stereo image and the end of the corresponding tracking process even though no pose is published in that moment (it could be published if needed). The tracking time comparison is shown in Figure~\ref{fig:times}. 

The fastest system is SVO 2.0 and this is explained by the fact that it is a sparse method that only extracts features (corners and edgelets) for keyframes and not for every frame as in most of the other systems. Once obtained, they are directly tracked from pixel intensities. This is the reason why it is a semi-direct approach. FLVIS also shows a good performance in terms of tracking times. Its speed is due to its estimation method that is based on feedback and feedforward loops and avoids a full probabilistic treatment of the measurements. Furthermore, it tracks the extracted features with KLT \cite{lucas1981iterative} avoiding description and matching stages. Among the slowest systems we find ROVIO, S-MSCKF and VINS-Fusion.

Considering these results and what is shown in Table~\ref{tab:aperpe} and Table~\ref{tab:orbaperpe}, ORB-SLAM3 offers the best trade-off between accuracy and computational load, thus being the most suitable localization approach for autonomous navigation in the countryside. Nevertheless, it should be pointed out that even including this system that best performed, none of the assessed offers an acceptable error. The distance between crop rows, in the soybean case for example, is approximately \SI{0.5}{\meter}. This means that in the best case, with the wheels starting in the middle, they would have a margin of \SI{0.25}{\meter} to each side. However, the minimum ATE obtained by ORB-SLAM3, in sequence 01, was \SI{0.86}{\meter}. This shows that the state-of-the-art VIO systems are not accurate enough (by themselves) for a reliable localization of our weed removal robot during its operation in the crop field.


\section{Conclusions}
\label{sec:conclusions}
In this work, we presented an evaluation of the most relevant state-of-the-art VIO/SLAM systems in an agricultural environment. In particular, we performed the evaluation on the Rosario Dataset, a collection of sensor data gathered by our weed removal robot in a soybean field.

Besides the evaluation itself, we also ported each assessed VIO system to a docker container and provided a set of scripts to make the results reproducible and to facilitate the study of the presented systems, even on other datasets.

The exhaustive evaluation also helped to detect some IMU time related issues caused during the gathering of the dataset and thus, this work contributes to the Rosario Dataset validation.

The analysis shows the limitations of the evaluated VIO systems on the agricultural field. Several aspects such as highly repetitive appearance of scenes intrinsic to agricultural fields, leaves movement caused by the wind and vibration caused by the rough terrain, considerably affect the performance of the systems. R-VIO and Kimera-VIO fail to initializes properly under robot movement, this is the case for the sequences 01 and 05 of the dataset. On the other hand, REBVO turns out to be sensitive to IMU saturation, which is an undesirable consequence of the irregular soil. State-of-the-art systems like VINS-Fusion, OKVIS and SVO 2.0 often fail to work properly in some sequences, proving to be insufficiently robust in agricultural fields. It should be noted that even though the results of S-MSCKF are acceptable, its high processing time is not suitable for real-time application. Lastly, ORB-SLAM3 achieves the lowest ATE in some sequences and a reasonable processing time, providing a good compromise between accuracy and computational cost. Nevertheless, this error is not small enough for navigation tasks in agricultural environments, where a high degree of accuracy is required.

Considering the results of this work, we select ORB-SLAM3 as an initial architecture to develop the localization system of the weed removal robot. In order to improve the robustness and accuracy of such system, we plan to include global measurements as the ones provided by standard GNSS and magnetometers as well as wheel odometry measurements and specific constraints related to the environment (e.g. roughly planar terrain). 

\section*{Acknowledgments}
\label{sec:acknowledgments}
This work is part of the \textit{Development of a weed removal mobile robot} project at CIFASIS (CONICET-UNR).
We want to thank the staff of the CCT-Rosario Computational Center, member of the High Performance Computing National System (SNCAD, MinCyT-Argentina), for their help with the data processing.

\bibliography{biblio}
\end{document}